\documentclass{article}

\usepackage{microtype}
\usepackage{graphicx}
\usepackage{subcaption}
\usepackage{booktabs} 

\usepackage{hyperref}

\usepackage[accepted]{icml2026}

\usepackage{amsfonts,bm,amscd}
\usepackage{mathrsfs}
\usepackage{amsmath}
\usepackage{amssymb}
\usepackage{mathtools}
\usepackage{amsthm}
\usepackage{pifont}
\usepackage{multirow}
\usepackage{colortbl}
\usepackage{multicol}
\usepackage{tabularray}
\usepackage{xcolor}
\usepackage{bbding}
\usepackage[capitalize,noabbrev]{cleveref}

\theoremstyle{plain}
\newtheorem{theorem}{Theorem}[section]

\newtheorem{lemma}[theorem]{Lemma}
\newtheorem{corollary}[theorem]{Corollary}
\theoremstyle{definition}

\theoremstyle{remark}
\newtheorem{remark}[theorem]{Remark}

\usepackage[textsize=tiny]{todonotes}

\begin{document}

\twocolumn[
  \icmltitle{Solving Prior Distribution Mismatch in Diffusion Models via Optimal Transport}



  \icmlsetsymbol{equal}{*}

  \begin{icmlauthorlist}
    \icmlauthor{Zhanpeng Wang}{yyy}
    \icmlauthor{Shenghao Li}{yyy}
    \icmlauthor{Jiameng Che}{yyy}
    \icmlauthor{Chen Wang}{yyy}
    \icmlauthor{SHANGLING JUI}{comp}
    \icmlauthor{Na Lei$^{\dagger}$}{yyy}
    \icmlauthor{Zhongxuan Luo}{yyy}
  \end{icmlauthorlist}

  \icmlaffiliation{yyy}{School of Software, Dalian University of Technology, Dalian, China}
  \icmlaffiliation{comp}{Lagrange Mathematics and Computing Research Center, Huawei, Paris, France}

  \icmlcorrespondingauthor{Na Lei$^{\dagger}$}{nalei@dlut.edu.cn}


  \vskip 0.3in
]



\printAffiliationsAndNotice{}  

\begin{abstract}
	Diffusion Models (DMs) have achieved remarkable progress in generative modeling. However, the mismatch between the forward terminal distribution and reverse initial distribution introduces prior error, leading to deviations of sampling trajectories from the true distribution and severely limiting model performance. This issue further triggers cascading problems, including non-zero Signal-to-Noise Ratio, accumulated denoising errors, degraded generation quality, and constrained sampling efficiency. To address this issue, this paper proposes a prior error elimination framework based on Optimal Transport (OT). Specifically, an OT map from the reverse initial distribution to the forward terminal distribution is constructed to achieve precise matching of the two distributions. Meanwhile, the upper bound of the prior error is quantified using the Wasserstein distance, proving that the prior error can be effectively eliminated via the OT map. Additionally, by deriving the asymptotic consistency between dynamic OT and probability flow, this method is revealed to be highly compatible with the intrinsic mechanism of the diffusion process. Experimental results demo-\\nstrate that the proposed method completely eliminates the prior error both theoretically and practically, providing a universal and rigorous solution for optimizing the performance of DMs.
\end{abstract}

\section{Introduction}
\begin{figure}[t]
	\setlength{\abovecaptionskip}{0cm}
	\setcounter{subfigure}{0}
	\centering
	\begin{subfigure}[b]{0.484\textwidth}
		\includegraphics[width = 0.989\textwidth]{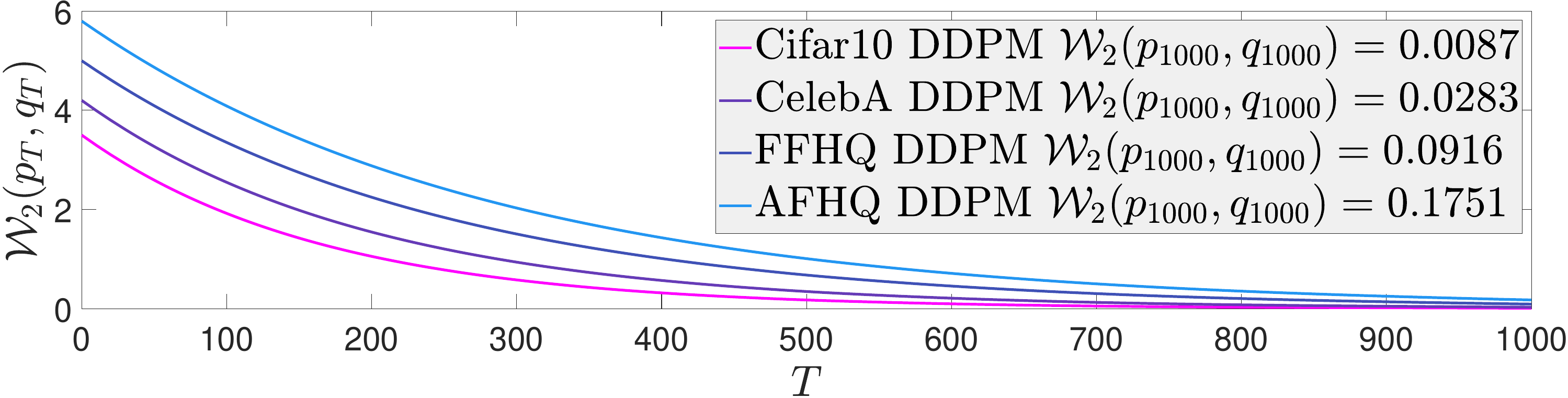}
		\caption{Variation of Prior Error with Termination Time $T$}
		\label{fig:W2_forward_reverse:a}
	\end{subfigure}
	\\
	\begin{subfigure}[b]{0.484\textwidth}
		\includegraphics[width = 0.989\textwidth]{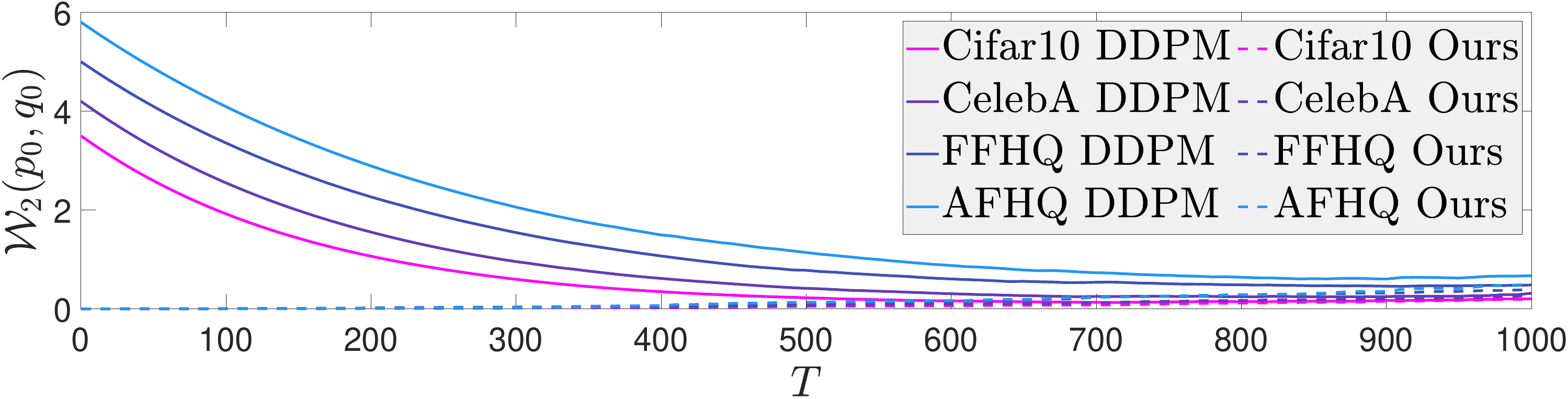}
		\caption{Impact of Prior Error on the Generative Distribution}
		\label{fig:W2_forward_reverse:b}
	\end{subfigure}
	\caption{In the forward process, DDPM \cite{ho2020denoising} corrupts the true data distribution $p_{0}$, directing it toward the Gaussian distribution $\mathcal{N}(\boldsymbol{0},\boldsymbol{I})$ but practically terminating at $p_{T}$. In the reverse process, it adopts $q_{T}=\mathcal{N}(\boldsymbol{0},\boldsymbol{I})$ as the prior, with this mismatch inducing the prior error $\mathcal{W}_{2}(p_{T},q_{T})$, which increases the Wasserstein difference between the generative distribution $q_{0}$ and $p_{0}$. To mitigate this adverse effect, DDPM can only increase diffusion time $T$, leading to elevated training costs, low sampling efficiency, and excessive error accumulation. In contrast, our method, which eliminates the prior error via OT map, is more robust to variations in $T$, thus enabling extension to accelerated sampling.}\label{fig:W2_forward_reverse}
\end{figure}

Diffusion Models (DMs) \cite{ho2020denoising,song2020improved,song2020score} have emerged as the core backbone of modern generative artificial intelligence, revolutionizing a range of tasks from image synthesis \cite{ho2020denoising,rombach2022high}, cross-domain translation \cite{su2022dual,zhao2022egsde} to text-to-3D generation \cite{poole2022dreamfusion,chen2023fantasia3d}. Their success stems from an elegant two-stage mechanism: the forward process gradually corrupts data into noise (Fig. \ref{fig:W2_forward_reverse:a}), while the reverse process reconstructs data from this noise via score matching \cite{song2020score}. However, this mechanism relies on a critical theoretical assumption that the forward terminal distribution must exactly match the reverse initial distribution (typically a standard Gaussian distribution $\mathcal{N}(\boldsymbol{0}, \boldsymbol{I})$). In practical applications, this assumption is rarely satisfied, thereby giving rise to the pervasive issue of prior distribution mismatch (Fig. \ref{fig:W2_forward_reverse:b}), which undermines the fundamental basis of DMs.
\vspace{-15pt}

Prior distribution mismatch manifests in various destructive forms. Most notably, it leads to a non-zero signal-to-noise ratio (SNR) at the final time step $T$ of the forward process \cite{lin2024common} (see Lemma D.1 for equivalence between the non-zero SNR problem and prior distribution mismatch): the terminal state $\boldsymbol{x}_T$ is not pure Gaussian noise, but retains low-frequency signals from the original data (e.g., $\boldsymbol{x}_T = 0.068265\boldsymbol{x}_0 + 0.997667\boldsymbol{\epsilon}$ in Stable Diffusion \cite{lin2024common,wang2025silent}), violating the initial distribution assumption of the reverse process. This discrepancy creates a training-inference gap: during the training phase, DMs learn to denoise from a non-zero SNR state with residual data cues; whereas during the inference phase, they are forced to start from pure Gaussian noise (zero SNR). This inconsistency in denoising logic leads to the accumulation of errors \cite{kohler2024imagine}. Consequently, generated samples suffer from limited dynamic range, reduced fidelity, and semantic distortion \cite{wu2024freeinit,yuan2025freqprior}. In cross-domain translation tasks, the mismatch between the latent distributions of source- and target-domain DMs exacerbates these issues, causing the translation trajectory to deviate from the true target distribution \cite{wang2025ot}. Even in single-domain generation tasks, prior mismatch compels the model to compensate for distribution discrepancies by increasing the number of diffusion steps, significantly raising computational costs \cite{li2025optimal}. Additionally, ODE-based accelerated sampling methods (e.g., DPM-Solver \cite{lu2022dpm}, DEIS \cite{zhang2023fast}, and UniPC \cite{zhao2023unipc}) do not account for prior errors,
leading to the accumulation of numerical errors under long-term diffusion. 

Existing mitigation strategies are mostly heuristic and fragmented. Some methods adjust the noise schedule to force a zero terminal SNR \cite{lin2024common,barrault2024large,grosskopf2025histdist}, but they overlook distribution mismatch within DMs and may even distort the geometric structure of data. Other approaches introduce ad-hoc correction terms or modify the variance of the reverse process \cite{wu2024freeinit,kohler2024imagine,zhang2024tackling,yuan2025freqprior, wang2025silent}, yet they lack theoretical guarantees and fail to fundamentally resolve the distribution alignment problem. Compared with the methods mentioned above, a more intuitive and general approach is to employ a generator $\mathcal{G}(\cdot)$ to align the probabilities between the forward terminal distribution and reverse initial distribution. While such approaches can not only bridge the gap caused by prior mismatch but also be easily extended to accelerated generation, they suffer from the following drawbacks (Tab. \ref{tab:Comparison of single-step generator operations}): (1) They only focus on improving sampling speed and generation quality, lacking in-depth exploration of the critical prior error theory; (2) Continuous $\mathcal{G}(\cdot)$ makes it impossible to eliminate prior error completely in practical application scenarios; (3) The constructed models tend to inherit the defects of $\mathcal{G}(\cdot)$, such as mode mixture \cite{li2025ambiguity}.
\begin{table}[htbp]
	\centering
	\caption{Generator-construction-type strategies related to prior mismatch. A brief introduction to these methods is provided in Appendix A.2. Notably, although \cite{li2023dpm} and \cite{li2025optimal} also involve OT, our work offers deeper theoretical insights compared to them, with detailed differences summarized in Tab. 4}
	\scalebox{0.655}{
	\begin{tabular}{c|cccc }
		\hline
		Methods&$\mathcal{G}(\cdot)$&Final effect&Theoretical& Diversity \\ \hline
		\begin{tabular}[c]{@{}c@{}}DDPM, DDIM,\\ ODE-based, etc\end{tabular} &Identity map&Ignore&\XSolidBrush&Mixture\\ \hline
		\cite{zheng2022truncated}&GANs/CT&Alleviate&\XSolidBrush&Mixture\\
		\cite{lyu2022accelerating}&GANs/VAEs&Alleviate&\XSolidBrush&Mixture\\
		\cite{Chung_2022_CVPR}&GANs&Alleviate&\XSolidBrush&Mixture\\
		\cite{li2023dpm}&OT&Eliminate&\XSolidBrush&Well\\
		\cite{franzese2023much}&DPGMM/Glow&Alleviate&\XSolidBrush&Mixture\\		
		\cite{wang2024optimizing}&Anchored  map&Alleviate&\XSolidBrush&Mixture\\		
		\cite{guo2024accelerating}&nnUNeT/AttUNet &Alleviate&\XSolidBrush&Mixture\\		
		\cite{zand2024diffusion}&Flow-based &Alleviate&\XSolidBrush&Mixture\\		
		\cite{song2024torch}&Prometheus SDEs &Alleviate&\XSolidBrush&Mixture\\
		\cite{everaert2024exploiting}&Anchored  map&Alleviate&\XSolidBrush&Mixture\\
		\cite{li2025optimal}&AE-OT&Alleviate&\XSolidBrush&Well\\
		\cite{liao2025diffusiondrive}&Anchored  map&Alleviate&\XSolidBrush&Mixture\\
		\cite{dong2025flow}&Flow-based&Alleviate&\XSolidBrush&Mixture\\
		Ours&OT&Eliminate&\CheckmarkBold&Well\\ \hline
	\end{tabular}}
    \label{tab:Comparison of single-step generator operations}
    \vspace{-14pt}
\end{table}

To address these limitations, we argue that prior distribution mismatch is a fundamental geometric problem, which necessitates a rigorous and unified solution. As a powerful tool for measuring and aligning probability distributions, Optimal Transport (OT) can identify the minimum-cost map between distributions. Furthermore, it inherently preserves the intrinsic structure of data, making it an ideal framework for resolving the prior distribution mismatch. Therefore, our main contributions and findings are summarized as follows:

$ \bullet $ We investigate the prior mismatch existing in DMs, and prove that this issue is equivalent to the non-zero SNR problem (Lemma D.1). This core contradiction creates a critical theoretical gap: specifically, even if DMs can accurately learn the score function (Theorem \ref{theorem:Wasserstein_distance_upper_bound_DM}), their final sampling trajectory still deviates from the true data distribution.

$ \bullet $ We introduce an OT-based prior error eliminator. Specifically, an OT map from the reverse initial distribution to the forward terminal distribution is constructed. Meanwhile, the error upper bound (Theorem \ref{corollary:Wasserstein_distance_upper_bound_OUR}) proves that our method eliminate prior error effectively. Furthermore, by deriving the asymptotic consistency between dynamic OT and probability flow (Theorem \ref{theorem:OT_contraction}), this method is shown to be highly compatible with the intrinsic mechanism of DMs.

$ \bullet $ Experimental results on multiple image datasets demonstrate that our method completely eliminates prior error both theoretically (Lemma \ref{lemma:eliminate the prior error}) and practically (Tab. \ref{tab:Comparison of prior error elimination effects among different single-step generators}), thereby improving generation quality while accelerating inference (Tab. \ref{tab:Quantitative comparison}). Overall, this paper fills the theoretical gap in the direction of prior error for DMs and provides a general, rigorous solution for optimizing the performance of DMs.

Notations, Definitions and Assumptions related to our theoretical contributions are elaborated in Appendix C.

\section{Preliminaries and Related Works}\label{sec2}
\subsection{Static/Dynamic Optimal Transport}
This section presents a succinct overview of static and dynamic OT problems under the quadratic cost $c(\boldsymbol{x},\boldsymbol{y})=\frac{1}{2}\|\boldsymbol{x}-\boldsymbol{y}\|_{2}^{2}$, along with their mutual compatibility.

Assuming $ \rho_{0},\rho_{1}\in\mathscr{P}(\mathbb{R}^{n},\mathcal{W}_{2}) $ are continuous probability density functions, map $ M:\mathbb{R}^{n}\to\mathbb{R}^{n} $ is called measure preserving (denoted $ M_{\#}\rho_{0}=\rho_{1} $) if it satisfies $ \int_{B}\rho_{1}(\boldsymbol{y})d\boldsymbol{y}=\int_{M^{-1}(B)}\rho_{0}(\boldsymbol{x})d\boldsymbol{x} $ for any Borel set $ B\subset\mathbb{R}^{n} $. Introduced in ref. \cite{monge1781memoire}, the Monge problem aims to find the OT map $ M_{ot} $ among all measure preserving maps, that is\vspace{-10pt}

{\setlength{\abovedisplayskip}{0.15pt}
\setlength{\belowdisplayskip}{1pt}
\small
\begin{equation}\label{eq:Monge_problem}
	M_{ot}=\arg\min\limits_{M_{\#}\rho_{0}=\rho_{1}}\frac{1}{2}\int_{\mathbb{R}^{n}}\| \boldsymbol{x}-M(\boldsymbol{x}) \|_{2}^{2}\rho_{0}(\boldsymbol{x}) d\boldsymbol{x},
\end{equation}
}\vspace{-15pt}

the one that minimizes the total transport cost. Moreover, there exists a convex Brenier potential $ u\left(\boldsymbol{x}\right):\mathbb{R}^{n}\to\mathbb{R} $ \cite{brenier1991polar} that satisfies that $ \nabla u=M_{ot} $ is the unique solution to \eqref{eq:Monge_problem} when quadratic cost. From Jacobian equation, one can get the necessary condition for Brenier potential, namely, Monge-Ampère equation \cite{villani2009optimal}\vspace{-10pt}

{\setlength{\abovedisplayskip}{0.15pt}
\setlength{\belowdisplayskip}{1pt}
\small
\begin{equation}\label{eq:Monge-Ampere_equation}
	\det \nabla^{2}u(\boldsymbol{x})=\frac{\rho_{0}(\boldsymbol{x})}{\rho_{1}\circ\nabla u(\boldsymbol{x})}.
\end{equation}
}\vspace{-15pt}

Benamou et al. \cite{benamou2000computational} proposed an analytical framework connecting \eqref{eq:Monge_problem} to continuum mechanics, analyzing the dynamic evolution of time-dependent transport from $\rho_0$ ($t=0$) to $\rho_1$ ($t=1$). The intermediate process $\rho_t$ ($t\in[0,1]$) satisfies the continuity equation\vspace{-10pt}

{\setlength{\abovedisplayskip}{0.15pt}
\setlength{\belowdisplayskip}{1pt}
\small
\begin{equation}\label{eq:continuity_equation}
	\frac{\partial \rho_{t}(\boldsymbol{x})}{\partial t}+\nabla\cdot(\boldsymbol{v}(\boldsymbol{x},t)\rho_{t}(\boldsymbol{x}))=0,
\end{equation}
}\vspace{-15pt}

where $\boldsymbol{v}(\boldsymbol{x},t):\mathbb{R}^{n}\times[0,1]\to\mathbb{R}^{n}$ is the time-dependent transport velocity field. The Benamou-Brenier formulation minimizes the total kinetic energy over velocity fields\vspace{-10pt}

{\setlength{\abovedisplayskip}{0.15pt}
\setlength{\belowdisplayskip}{1pt}
\small
\begin{equation}\label{eq:W_2_welocity_field}
	\boldsymbol{v}_{ot}(\boldsymbol{x},t)=\arg\min_{\boldsymbol{v}}\frac{1}{2}\int_{0}^{1}\mathbb{E}_{\rho_{t}}[\|\boldsymbol{v}(\boldsymbol{x},t)\|_{2}^{2}]dt
\end{equation}
}\vspace{-15pt}

satisfying $\rho_{0},\rho_{1}$ (boundary conditions) and \eqref{eq:continuity_equation} (mass conservation). The static Monge \eqref{eq:Monge_problem} and dynamic Benamou-Brenier problem \eqref{eq:W_2_welocity_field} exhibit equivalence and compatibility in terms of the quadratic transport cost \cite{villani2009optimal}. Specifically, for identical boundary conditions, the total tran-sport cost of the OT map $M_{ot}$ is equal to the total kinetic energy of the optimal velocity field $\boldsymbol{v}_{ot}$, so they are both taken as $\mathcal{W}_{2}$ metrics in (26). In addition, they are convertible via McCann displacement interpolation \cite{mccann1997convexity}.
\subsection{Diffusion Models and Probability Flow}
The intrinsic driver of DMs is stochastic differential equation (SDE), which is usually expressed as\vspace{-10pt}

{\setlength{\abovedisplayskip}{0.15pt}
\setlength{\belowdisplayskip}{1pt}
\small
\begin{equation}\label{eq:SDE}
	d\boldsymbol{x}=-f(t)\boldsymbol{x}dt+g(t)d\boldsymbol{W}_{t}, \boldsymbol{x}(0)\sim p_{0}(\boldsymbol{x}),
\end{equation}
}\vspace{-15pt}

where $ f(t):[0,T]\to\mathbb{R}_{\geq 0} $ means the non-negative drift coefficient, $ g(t):[0,T]\to\mathbb{R}_{> 0} $ is positive diffusion term, $ p_{0} $ is the initial distribution and $ \boldsymbol{W}_{t} $ means a time-dependent $ n $-dimensional standard Wiener process. The marginal distribution $ p_{t} $ of the solution to \eqref{eq:SDE} at time $ t $ follows the Fokker-Planck equation (FPE) \cite{risken1996fokker}\vspace{-10pt}

{\setlength{\abovedisplayskip}{0.15pt}
\setlength{\belowdisplayskip}{1pt}
\small
\begin{equation}\label{eq:Fokker_Planck_equation}
	\frac{\partial p_{t}(\boldsymbol{x})}{\partial t}+\nabla\cdot(-f(t)\boldsymbol{x}p_{t}(\boldsymbol{x}))-\frac{g(t)^{2}}{2}\Delta p_{t}(\boldsymbol{x})=0,
\end{equation}
}\vspace{-15pt}

and admits a unique steady-state Gibbs distribution $p_{\infty}$ \cite{gardiner1985handbook}. According to \eqref{eq:Fokker_Planck_equation}, there exists a deterministic process, which we call the probability flow ordinary differential equation (ODE) \cite{song2020score}\vspace{-10pt}

{\setlength{\abovedisplayskip}{0.15pt}
\setlength{\belowdisplayskip}{1pt}
\small
\begin{equation}\label{eq:probability_flow_ODE}
	\frac{d\boldsymbol{x}}{dt}=-f(t)\boldsymbol{x}-\frac{g(t)^{2}}{2}\nabla\log p_{t}(\boldsymbol{x}),\boldsymbol{x}(0)\sim p_{0}(\boldsymbol{x})
\end{equation}
}\vspace{-15pt}

that shares the same marginal probability density $ p_{t} $ with \eqref{eq:SDE}. For DMs training, a parametric neural network $\boldsymbol{S}_{\boldsymbol{\theta}}(\boldsymbol{x},t)$ approximates $\nabla\log p_{t}(\boldsymbol{x})$ \cite{song2020score}, and the target loss function $\mathcal{J}_{SM}(\boldsymbol{\theta},\phi,0,T)$ is the weighted MSE\vspace{-8pt}

{\setlength{\abovedisplayskip}{0.15pt}
\setlength{\belowdisplayskip}{1pt}
\small
\begin{equation}\label{eq:mse_loss_function}
	\mathcal{J}_{SM}=\frac{1}{2}\int_{0}^{T}\phi(t)\mathbb{E}_{p_{t}}[ \| \boldsymbol{S}_{\boldsymbol{\theta}}(\boldsymbol{x},t)-\nabla\log p_{t}(\boldsymbol{x}) \|_{2}^{2}]dt,
\end{equation}
}\vspace{-8pt}

where $ \boldsymbol{\theta} $ is a learnable parameter and $ \phi\left(t\right):\left[0,T\right]\to\mathbb{R}_{> 0} $ means positive time-dependent weight. In practice, however, $\mathcal{J}_{SM}(\boldsymbol{\theta},\phi,0,T)$ is often difficult to quantify due to lack of information about $\nabla\log p_{t}(\boldsymbol{x})$ \cite{song2021maximum}. Therefore, it is common to convert it to a treatable conditional score matching loss $\mathcal{J}_{DSM}(\boldsymbol{\theta},\phi,0,T)$ \cite{vincent2011connection} via letting $p_{t}(\boldsymbol{x})=p_{t}(\boldsymbol{x}(0))p_{0t}(\boldsymbol{x}|\boldsymbol{x}(0))$, here $p_{0t}$ indicates the probabilistic transition kernel from time $0$ to $t$.
\subsection{Related Works}\label{sec:related works}
\textbf{Intrinsic Relationship Between DMs and OT.} The authors of \cite{khrulkov2022understanding}
show that the probability flow can be considered as an OT map under the initial assumption of a multivariate normal distribution. However, a counterexample of continuous initial distribution is provided in ref. \cite{lavenant2022flow} to show that the probability flow is not optimal. Recently, Zhang et al. \cite{zhang2024formulating} further extend their conclusions, and argue that the probability flow over any closed interval in $ (0,\infty) $ coincides with the OT map under the condition of finite training samples. In addition, Kwon et al. \cite{kwon2022score} indicate that DMs also minimize the Wasserstein distance between the true distribution and the generated distribution.

\noindent\textbf{Singularity of DMs.} Under the initial condition of discrete probability density, the score $ \nabla\log p_{t}\left(\boldsymbol{x}\right) $ does not exist at $ t=0 $, resulting in $ \boldsymbol{S}_{\boldsymbol{\theta}}\left(\boldsymbol{x},t\right) $ cannot be Lipschitz continuous across $ 
[0,T] $ \cite{zhang2024tackling}. Therefore, Song et al. \cite{song2020score} suggest to choose a small $ \varepsilon>0 $ to terminate the sampling process to ensure the quality of the generated image \cite{lu2022dpm,song2023consistency,zhang2024tackling,lin2024schedule}. In addition, there are numerous theoretical analysis based on this idea \cite{benton2023linear,yang2023lipschitz,li2024good}.

For the relevant research pertaining to \textit{Non-Zero SNR in DMs} and \textit{Generator Mitigation Strategy for Prior Error in DMs}, please refer to Appendix A.
\section{OT-Based Prior Error Eliminator}\label{sec3}
In this section, we outline an OT-grounded solution to resolve prior distribution mismatch in DMs. After mathematically analyzing the method for theoretical justification, we apply this insight to boost the generative efficiency of DMs.
\subsection{Methodology}
To address the prior error within DMs, we propose computing OT map $\nabla u_{T}^{\gets}$, which maps steady-state distribution $p_{\infty}$ to terminal distribution $p_{T}$ of the forward process. We then use $q_{T} = \nabla u_{T}^{\gets}(p_{\infty})$ as initial distribution for the reverse process. Notably, rectifying the prior distribution mismatch via OT significantly enhances the consistency between the generated distribution $q_{0}$ and true distribution $p_{0}$ (Fig. \ref{fig:W2_forward_reverse:b}). Detailed error analysis is shown in Theorem \ref{corollary:Wasserstein_distance_upper_bound_OUR}.

To compute the OT map \(\nabla u_{T}^{\gets}\) more precisely, we introduce a geometric perspective to solve the Monge-Ampère equation \eqref{eq:Monge-Ampere_equation}. The geometric variational method—established by \cite{gu2016variational} and widely applied to deep learning via convex optimization \cite{lei2019geometric,an2019ae}—reveals the intrinsic connection between Brenier \cite{brenier1991polar} and Alexandrov theory. We briefly present it in the discrete setting, with natural generalization to the continuous case when the number of samples is sufficiently large.

Since probability density $p_{T},p_{\infty}$ decay exponentially when sufficiently far from the coordinate origin and satisfy the \textit{compact-like property} (Proposition C.14), we can select a closed set $\Omega\subset\mathbb{R}^{n}$ as the source domain. Given discrete dataset $ \mathcal{X}_{T}=\{\boldsymbol{x}_{T}^{i}\}_{i\in\mathcal{I}}\overset{i.i.d}{\sim}p_{T} $, the target Dirac probability density is $ p_{\mathcal{X}_{T}}=\frac{1}{|\mathcal{I}|}\sum_{i\in\mathcal{I}}\delta ( \boldsymbol{x}-\boldsymbol{x}_{T}^{i} ) $. Gu et al. construct a cluster of optimal hyperplanes $ \{\pi_{i}(\boldsymbol{z})=\langle \boldsymbol{z}, \boldsymbol{x}_{T}^{i} \rangle +h_{i}\}_{i\in\mathcal{I}} $ and denote their upper envelope as $ env(\{\pi_{i}\}_{i\in\mathcal{I}}) $ \cite{gu2016variational}, which corresponds to the graph of the convex piecewise linear Brenier function\vspace{-10pt}

{\setlength{\abovedisplayskip}{0.15pt}
\setlength{\belowdisplayskip}{1pt}
\small
\begin{equation}\label{eq:h parameterized brenier}
	u_{T}^{\gets}(\boldsymbol{z})=\max_{i\in\mathcal{I}}\{\langle \boldsymbol{z}, \boldsymbol{x}_{T}^{i} \rangle +h_{i}\},\boldsymbol{z}\in\Omega.
\end{equation}
}\vspace{-15pt}

Notably, as $|\mathcal{I}|\to\infty$, the semi-discrete OT $\nabla u_{T}^{\gets}$ in \eqref{eq:h parameterized brenier} converges to continuous case. Under the above formulation, the energy functional $ E\left(\varphi\right) $ in (79) can be rewritten as\vspace{-10pt}

{\setlength{\abovedisplayskip}{0.15pt}
\setlength{\belowdisplayskip}{1pt}
\small
\begin{equation}\label{eq:energy E(h)}
	E(\boldsymbol{h})=\frac{1}{|\mathcal{I}|}\sum_{i\in\mathcal{I}}h_{i} - \int_{0}^{\boldsymbol{h}} \sum_{i\in\mathcal{I}} w_{i}(\eta) d\eta_{i},
\end{equation}
}\vspace{-10pt}

where $w_{i}(\eta)=\int_{W^{i}_{T}}p_{\infty}(\boldsymbol{z})d\boldsymbol{z}$, $\boldsymbol{h}=(h_{1},h_{2},\cdots,h_{|\mathcal{I}|})^{T}$ and $W_{T}^{i}=\{\boldsymbol{z}\in\Omega|\langle \boldsymbol{z}, \boldsymbol{x}_{T}^{i}-\boldsymbol{x}_{T}^{j} \rangle \geq h_{j}-h_{i},\forall j\in\mathcal{I}\}$, within the admissible potential space\vspace{-10pt}

{\setlength{\abovedisplayskip}{0.15pt}
\setlength{\belowdisplayskip}{1pt}
\small
\begin{equation}\label{eq:admissible potential space}
	\mathcal{A}=\{\boldsymbol{h}|w_{i}(\boldsymbol{h})>0,i\in\mathcal{I}\}\cap\{\boldsymbol{h}|\sum_{i\in\mathcal{I}}h_{i}=1\}.
\end{equation}
}\vspace{-15pt}

We maximize $E(\boldsymbol{h})$ in \eqref{eq:energy E(h)} to obtain the optimal $\boldsymbol{h}$, which is then substituted into \eqref{eq:h parameterized brenier} to yield the approximate convex piecewise linear Brenier potential $u_{\boldsymbol{h}}^\gets$. The semi-discrete OT map $\nabla u_{\boldsymbol{h}}^\gets$ induces a cell decomposition $\Omega=\bigcup_{i\in\mathcal{I}} W_T^i$, where each cell satisfies $\nabla u_{\boldsymbol{h}}^\gets(W_T^i)=\boldsymbol{x}_T^i$. The complete process of computing OT map is organized in Algorithm \ref{alg:Computing_OT}.
\begin{algorithm}[htbp]
	\footnotesize
	\caption{{\small Computing OT Map from $p_{\infty}$ to $p_{\mathcal{X}_{T}}$ via Geometric Variation Principle}}
	\label{alg:Computing_OT}
	\textbf{Input}: Data distribution $ p_{0} $, diffusion termination time $T$, discrete time interval $ \Delta t $ such that $T=m\Delta t$, number of Monte Carlo samples $N_{mc}$, positive integer $s$.
	\begin{algorithmic}[1]
		\STATE $\mathcal{X}_{0}\gets\{\boldsymbol{x}_{0}^{i}\}_{i\in\mathcal{I}}\overset{i.i.d}{\sim}p_{0}$.\\
		\FOR{$ i=0:m-1 $}
		\STATE $ \mathcal{X}_{t_{i+1}}\gets \mathcal{X}_{t_{i}}-f(t_{i})\mathcal{X}_{t_{i}}\Delta t+g(t_{i})\sqrt{\Delta t}\boldsymbol{\xi}$, where $t_{i}=i\Delta t,\boldsymbol{\xi}\sim\mathcal{N}(\boldsymbol{0},\boldsymbol{I})$.
		\ENDFOR
		\STATE $\boldsymbol{h}\gets \boldsymbol{0}$.\\
		\REPEAT
		\STATE $w_{i}(\boldsymbol{h})\gets \frac{|\{\boldsymbol{z}|\boldsymbol{z}\in \{\boldsymbol{z}_{j}\}_{j=1}^{N}\cap W_{T}^{i}\}|}{N_{mc}}$, $\{\boldsymbol{z}_{j}\}_{j=1}^{N_{mc}}\overset{i.i.d}{\sim}p_{\infty}$,  $i\in\mathcal{I}$.\\
		\STATE $\nabla E(\boldsymbol{h})\gets(w_{1}(\boldsymbol{h})-\frac{1}{|\mathcal{I}|},\cdots,w_{|\mathcal{I}|}(\boldsymbol{h})-\frac{1}{|\mathcal{I}|})^{T}$.\\
		\STATE $\nabla E(\boldsymbol{h})\gets\nabla E(\boldsymbol{h})-mean(\nabla E(\boldsymbol{h}))$.\\
		\STATE Update $\boldsymbol{h}$ by Adam algorithm ($\beta_{1}=0.9$, $\beta_{2}=0.5$).\\
		\STATE $N_{mc}\gets N_{mc}\times 2$ if $E(\boldsymbol{h})$ has not decreased for $s$ steps.
		\UNTIL{Converge}
		\STATE $u_{\boldsymbol{h}}^{\gets}(\cdot)\gets\max\limits_{i\in\mathcal{I}}\{\langle\cdot, \boldsymbol{x}_{T}^{i} \rangle +h_{i}\}$.
	\end{algorithmic}
	\textbf{Return}: OT map $ \nabla u_{\boldsymbol{h}}^{\gets} $.
\end{algorithm}
\begin{remark}
	We adopt the Euler-Maruyama \cite{higham2001algorithmic} discretization for SDEs in Algorithms \ref{alg:Computing_OT}-\ref{alg:unconditional_generator}. Other SDE or ODE numerical schemes are also available.
\end{remark}

\subsection{Theoretical Analysis}
This section describes the main theoretical results. For more detailed analyses and proofs, refer to the Appendices D-E.

\textbf{Prior Distribution Mismatch.} In practical applications, the simple Gibbs distribution $ p_{\infty} $ instead of $ p_{T} $ is taken as the vanilla distribution $ q_{T} $ of the reverse process \cite{ho2020denoising,song2020score}, i.e., $ q_{T}=p_{\infty} $, so there exist a prior error $ \mathcal{W}_{2}\left(p_{T},q_{T}\right)\ne 0 $. We still have $ \mathcal{W}_{2}(p_{\varepsilon},q_{\varepsilon})\ne 0 $ in the light of Theorem \ref{theorem:Wasserstein_distance_upper_bound_DM} even if $ \boldsymbol{S}_{\boldsymbol{\theta}}\left(\boldsymbol{x},t\right) $ reaches precisely the optimal target $ \nabla\log p_{t}\left(\boldsymbol{x}\right) $ on $ \mathbb{R}^{n}\times\left[\varepsilon,T\right] $. In other words, the irreducibility of the potential gap makes the sampling process deviate in any way.
\begin{theorem}[Proof in Appendix E.4]\label{theorem:Wasserstein_distance_upper_bound_DM}
	If $ q_{T}=p_{\infty} $ and $ \boldsymbol{S}_{\boldsymbol{\theta}}\left(\boldsymbol{x},t\right)\equiv\nabla\log p_{t}\left(\boldsymbol{x}\right) $ on $ \mathbb{R}^{n}\times[\varepsilon,T] $, let $q_\varepsilon$ denote the distribution generated by DMs. Then $\mathcal{W}_2(p_\varepsilon, q_\varepsilon)$ admits the following estimation\vspace{-10pt}
	
	{\setlength{\abovedisplayskip}{0.15pt}
	\setlength{\belowdisplayskip}{1pt}
	\small
	\begin{equation}\label{eq:Wasserstein_distance_upper_bound_DM}
		\frac{\bar{I}(T)}{\bar{I}(\varepsilon)}\mathcal{W}_{2}(p_{T},q_{T})\leq\mathcal{W}_{2}(p_{\varepsilon},q_{\varepsilon})\leq\frac{I(T)}{I(\varepsilon)}\mathcal{W}_{2}(p_{T},q_{T}),
	\end{equation}
    }\vspace{-10pt}
    
    where $I(t)=\exp\left(\int_{0}^{t}(f(\tau)+g(\tau)^{2}L_{\boldsymbol{S}_{\boldsymbol{\theta}}}(\tau))d\tau\right)$, $\bar{I}(t)=\exp\left(\frac{1}{2}\int_{0}^{t}f(\tau)d\tau\right)$ are integrating factors, $L_{\boldsymbol{S}_{\boldsymbol{\theta}}}(t)$ is a continuous Lipschitz constant satisfying (28). 
\end{theorem}
\begin{remark}
	Theorem \ref{theorem:Wasserstein_distance_upper_bound_DM} implies that the reverse process derived via minimizing \eqref{eq:mse_loss_function} is necessarily associated with a non-vanishing deviation, provided that $\mathcal{W}_{2}(p_{T},q_{T})\ne 0$.
\end{remark}
However, the approach we propose effectively addresses this issue, and the following error analysis will demonstrate the validity of our method in resolving such deviations.

\textbf{Error Analysis of Our Method.} Let $u_{T}^{\gets}$ denote the Brenier potential satisfying the Monge-Ampère equation \eqref{eq:Monge-Ampere_equation} with boundary conditions $\rho_0 = p_\infty$ and $\rho_1 = p_T$; Algorithm \ref{alg:Computing_OT} enables us to approximate it via an $\boldsymbol{h}$-parameterized convex continuous network $u_{\boldsymbol{h}}^{\gets}$, yielding a surrogate $\nabla u_{\boldsymbol{h}}^{\gets}$ for the true OT map $\nabla u_{T}^{\gets}$, with Lemma D.5 providing an upper bound on the error between these two maps. Leveraging this OT map error bound, we further analyze the generative distribution error of DMs equipped with our OT-based prior error eliminator, leading to Theorem \ref{corollary:Wasserstein_distance_upper_bound_OUR} below.
\begin{theorem}[Proof in Appendix E.6]\label{corollary:Wasserstein_distance_upper_bound_OUR}
	The error upper bound of the generation distribution of DMs with OT prior eliminator is\vspace{-10pt}
	
	{\setlength{\abovedisplayskip}{0.15pt}
	\setlength{\belowdisplayskip}{1pt}
	\small
	\begin{equation}
		\mathcal{W}_{2}(p_{\varepsilon},q_{\varepsilon})\leq k_{1}\mathcal{J}_{SM}(\boldsymbol{\theta},\phi,\varepsilon,T)^{\frac{1}{2}}+k_{2}\left \| u_{T}^{\gets}-u_{\boldsymbol{h}}^{\gets} \right \|_{\infty}^{\frac{1}{2}},
	\end{equation}
	}\vspace{-10pt}
	
	here $k_{1}=\frac{\sqrt{2(T-\varepsilon)}}{I(\varepsilon)}$, $k_{2}=2^{\frac{8-n}{4}}\pi^{-\frac{n}{4}}K_\Omega K_{\mathcal{X}_T}\frac{I\left(T\right)}{I\left(\varepsilon\right)}$, $ \phi\left(t\right)=g\left(t\right)^{4}I\left(t\right)^{2} $, $\|\cdot\|_{\infty}$ is the infinite norm defined in (25), $K_{\Omega}$ is related only to $\Omega$, and $K_{\mathcal{X}_T}$ only to $\mathcal{X}_T$.
\end{theorem}
These theoretical insights rely on the idealized continuous-distribution assumption. In practice, however, we approximate continuous distributions with discrete samples—so we address the discrete case: it is less general than the continuous counterpart, but far more practically relevant. Notably, for discrete initial data, the dynamic OT map and probability flow exhibit consistency. We further show that DMs’ training/sampling correspond to two successive phases in dynamic OT map computation. Therefore, the OT map serves as the most intrinsic prior error eliminator for DMs.

\textbf{Probabilistic Flow in Discrete Setting.} For initial discrete probability density $ p_{0}(\boldsymbol{x})=\frac{1}{|\mathcal{I}|}\sum_{i\in\mathcal{I}}\delta(\boldsymbol{x}-\boldsymbol{x}_{0}^{i}),|\mathcal{I}|<\infty $, Zhang et al. \cite{zhang2024formulating} have proven that the probability flow, which serves as the analytical solution to \eqref{eq:probability_flow_ODE}, adheres to dynamic OT map over any closed time interval within $ (0,\infty) $. So $ \boldsymbol{v}_{ot}(\boldsymbol{x},t)=-f(t)\boldsymbol{x}-\frac{g(t)^{2}}{2}\nabla\log p_{t}(\boldsymbol{x}) $ is the corresponding optimal velocity field with the lowest kinetic energy under the $ \mathcal{W}_{2} $ metric. We define the probability flow from time $ s $ to $ t $ as $ M_{ot}^{s,t}:\mathbb{R}^{n}\to\mathbb{R}^{n} $, $ M_{ot}^{s,t}(\boldsymbol{x}_{s})=\boldsymbol{x}_{t} $, here $0<s,t<\infty$. It is worth noting that the probability flow is reversible \cite{lavenant2022flow}, and $ (M_{ot}^{s,t})^{-1}=M_{ot}^{t,s} $. From this perspective, the training and sampling processes of DMs are tantamount to distinct stages in computing time-dependent OT.

\textbf{(I)} Training stage: Matching the optimal velocity field. If only optimization results are considered, then the weighted MSE loss $\mathcal{J}_{SM}(\boldsymbol{\theta},\phi,\varepsilon,T)$ in \eqref{eq:mse_loss_function} over $ \left[\varepsilon,T\right] $ is equal to\vspace{-10pt}

{\setlength{\abovedisplayskip}{0.15pt}
\setlength{\belowdisplayskip}{1pt}
\small
\begin{equation}\label{eq:matching_optimal_velocity_field}
	\mathcal{J}_{SM}\iff\int_{\varepsilon}^{T}\phi(t)\mathbb{E}_{p_{t}}[ \| \boldsymbol{v}_{\boldsymbol{\theta}}(\boldsymbol{x},t)-\boldsymbol{v}_{\boldsymbol{ot}}(\boldsymbol{x},t)  \|_{2}^{2}]dt,
\end{equation}}\vspace{-10pt}

where $ \boldsymbol{v}_{\boldsymbol{\theta}}(\boldsymbol{x},t)=-f(t)\boldsymbol{x}-\frac{g(t)^{2}}{2}\boldsymbol{S}_{\boldsymbol{\theta}}(\boldsymbol{x},t) $.

\textbf{(II)} Sampling stage: Solving the probability flow ODE. Substituting the learned $ \boldsymbol{v}_{\boldsymbol{\theta}}(\boldsymbol{x},t) $ in \eqref{eq:matching_optimal_velocity_field} into \eqref{eq:probability_flow_ODE} gives an approximate reverse probability flow ODE\vspace{-10pt}

{\setlength{\abovedisplayskip}{0.15pt}
\setlength{\belowdisplayskip}{1pt}
\small
\begin{equation}\label{eq:approximation_probability_flow_ODE}
	d\boldsymbol{x}=-\boldsymbol{v}_{\boldsymbol{\theta}}(\boldsymbol{x},t)dt,t\in[\varepsilon,T],\boldsymbol{x}(T)\sim q_{T}(\boldsymbol{x}),
\end{equation}
}\vspace{-20pt}

which is locus equation of the sampling process with $ q_{T} $ as prior distribution. The solution of \eqref{eq:approximation_probability_flow_ODE} provides an approximation $ M_{\boldsymbol{\theta}}^{T,\varepsilon}:\mathbb{R}^{n}\to\mathbb{R}^{n} $ of dynamic OT map $ M_{ot}^{T,\varepsilon} $.

We first state Lemma \ref{lemma:eliminate the prior error}, which shows our method eliminates fully the prior error under discrete initial conditions—\\unlike other generators, which only mitigate it.
\begin{lemma}[Proof in Appendix E.7]\label{lemma:eliminate the prior error}
	Given the initial discrete data points $\{\boldsymbol{x}_{0}^{i}\}_{i\in\mathcal{I}}$, whose image at time $T$ under the probability flow is $\mathcal{X}_{T}=\{\boldsymbol{x}_{T}^{i}\}_{i\in\mathcal{I}}$, the OT map $\nabla u_{\boldsymbol{h}}^\gets$ is obtained via Algorithm \ref{alg:Computing_OT}. Then, provided that $\boldsymbol{h}$ lies within the admissible potential space $\mathcal{A}$ defined in \eqref{eq:admissible potential space}, the condition $\mathcal{W}_2(p_{\mathcal{X}_{T}}, \nabla u_{\boldsymbol{h}}^{\gets}(p_{\infty})) = 0$ is always satisfied upon removing duplicate points.
\end{lemma}
Drawing on the above analysis, Theorem \ref{theorem:OT_upper_bound_DM} and Corollaries \ref{corollary:OT_upper_bound_DM}–\ref{corollary:OT_upper_bound_DM2} bridge the gap between dynamic OT’s computational and DMs' training error. These insights are critical for characterizing the performance bounds of our method.
\begin{theorem}[Proof in Appendix E.8]\label{theorem:OT_upper_bound_DM}
	Let $ M_{ot}^{T,\varepsilon} $, $ M_{\boldsymbol{\theta}}^{T,\varepsilon} $ be analytical solutions to reverse probability flow ODE \eqref{eq:probability_flow_ODE} and \eqref{eq:approximation_probability_flow_ODE}, respectively, sharing initial condition $q_T = p_T$. Then\vspace{-10pt}
	
	{\setlength{\abovedisplayskip}{0.15pt}
	\setlength{\belowdisplayskip}{1pt}
	\small
	\begin{equation}
		\| M_{ot}^{T,\varepsilon}-M_{\boldsymbol{\theta}}^{T,\varepsilon} \|_{L_{2}\left(q_{T}\right)}\leq\sqrt{\frac{T-\varepsilon}{2\tilde{I}\left(\varepsilon\right)^{2}}}\mathcal{J}_{SM}(\boldsymbol{\theta},\phi,\varepsilon,T)^{\frac{1}{2}},
	\end{equation}
	}\vspace{-10pt}
	
	where $ \phi\left(t\right)=g\left(t\right)^{4}\tilde{I}\left(t\right)^{2} $ and $\tilde{I}\left(t\right)=\bar{I}\left(t\right)\sqrt{I\left(t\right)}$.
\end{theorem}
We give the following corollary directly from Theorem \ref{theorem:OT_upper_bound_DM}.
\begin{corollary}[Proof in Appendix E.9]\label{corollary:OT_upper_bound_DM}
	Under the same setting as Theorem \ref{theorem:OT_upper_bound_DM}. If $ p_{0t} $ fulfills variance condition $ \mathrm{Var} [\mathbb{E}[ ( \nabla\log p_{0t} ( \boldsymbol{x}|\boldsymbol{x}_{0}  )  )^{\top}|\boldsymbol{x}_{0}]]=0 $, then\vspace{-10pt}
	
	{\setlength{\abovedisplayskip}{0.15pt}
	\setlength{\belowdisplayskip}{1pt}
	\small
	\begin{equation}
		\| M_{ot}^{T,\varepsilon}-M_{\boldsymbol{\theta}}^{T,\varepsilon} \|_{L_{2}\left(q_{T}\right)}\leq\sqrt{\frac{T-\varepsilon}{2\tilde{I}\left(\varepsilon\right)^{2}}}\mathcal{J}_{DSM}(\boldsymbol{\theta},\phi,\varepsilon,T)^{\frac{1}{2}}.
	\end{equation}}
\end{corollary}
\begin{remark}
	Theorem \ref{theorem:OT_upper_bound_DM} and Corollary \ref{corollary:OT_upper_bound_DM} reveal collectively that DMs' training implicitly promotes parameterized transport map to converge toward the optimal ground truth.
\end{remark}
\begin{remark}
	DDPMs \cite{ho2020denoising} satisfy the variance condition in Corollary \ref{corollary:OT_upper_bound_DM} refer to \cite{kwon2022score}.
\end{remark}
By incorporating OT and pushforward of probability measures into Corollary \ref{corollary:OT_upper_bound_DM}, we derive Corollary \ref{corollary:OT_upper_bound_DM2}.
{\setlength{\abovedisplayskip}{0.15pt}
\setlength{\belowdisplayskip}{1pt}
\begin{corollary}[Proof in Appendix E.10]\label{corollary:OT_upper_bound_DM2}
	Under the same setting as Corollary \ref{corollary:OT_upper_bound_DM}. Let $ q_{\varepsilon}=(M_{\boldsymbol{\theta}}^{T,\varepsilon})_{\#}q_{T} $, if $ \boldsymbol{v}_{\boldsymbol{\theta}}\left(\boldsymbol{x},t\right) $ solves \eqref{eq:W_2_welocity_field} with boundary conditions $ \rho_{0}=q_{T} $, $ \rho_{1}=q_{\varepsilon} $, then\vspace{-10pt}
	
	{\setlength{\abovedisplayskip}{0.15pt}
	\setlength{\belowdisplayskip}{1pt}
	\small
	\begin{equation}
		\mathcal{W}_{2}\left(p_{\varepsilon},q_{\varepsilon}\right)\leq\sqrt{\frac{T-\varepsilon}{2\tilde{I}\left(\varepsilon\right)^{2}}}\mathcal{J}_{DSM}(\boldsymbol{\theta},\phi,\varepsilon,T)^{\frac{1}{2}}.
	\end{equation}}\vspace{-15pt}
\end{corollary}}
Building further on above discussions and contraction property of the Wasserstein distance \cite{carrillo2006contractions}, we analyze the compatibility of the OT problem \eqref{eq:Monge_problem} and \eqref{eq:W_2_welocity_field}, and establish that the probability flow exponentially converges to the gradient of the Monge-Ampère equation’s analytical solution over time. Specifically, we take $\nabla u_{T}^{\to}$ as a static OT map from $p_T$ to $p_\infty$, which is not the probability flow on the finite interval $[T,s]$ ($s>T$). However, Theorem \ref{theorem:OT_contraction} gives an upper bound of the difference between them and Remark \ref{remark:OT_contraction} states that this dissimilarity will exponentially approach to zero as $s$ increases. This remark also identifies $\nabla u_{T}^{\to}$ as the generalized probability flow on $[T,\infty)$.
\begin{theorem}[Proof in Appendix E.11]\label{theorem:OT_contraction}
	Under Assumption C.13, suppose $ u_{T}^{\to} $ is the analytical solution of the Monge-Ampère equation \eqref{eq:Monge-Ampere_equation} with boundary conditions $ \rho_{0}=p_{T} $ and $ \rho_{1}=p_{\infty} $, then there exists a positive constant $ K $ independent of $ s $ that satisfies\vspace{-10pt}
	
	{\setlength{\abovedisplayskip}{0.15pt}
	\setlength{\belowdisplayskip}{1pt}
	\small
	\begin{equation}\label{eq:OT_contraction}
		\| \nabla u_{T}^{\to}-M_{ot}^{T,s} \|_{L_{2}(p_{T})}\leq K\{\frac{\bar{I}\left(T\right)}{\bar{I}\left(s\right)}\mathcal{W}_{2}\left(p_{\infty},p_{T}\right)\}^{\frac{2}{15}}.
	\end{equation}}\vspace{-15pt}
\end{theorem}

Building on Theorem \ref{theorem:OT_contraction}, the asymptotic property as $s \to \infty$ is further detailed in Remark \ref{remark:OT_contraction}.
\begin{remark}\label{remark:OT_contraction}
	Since $f(t)$ makes $ \lim\limits_{s\to\infty}\int_{T}^{s}f\left(t\right)dt=\infty $ hold, we have $\lim\limits_{s\to\infty}\frac{\bar{I}\left(T\right)}{\bar{I}\left(s\right)}=0$ in \eqref{eq:OT_contraction}.
\end{remark}
Moreover, for the inverse map and Monge-Ampère equation under swapped boundary conditions, we state Remark \ref{remark:OT_contraction2}.
\begin{remark}\label{remark:OT_contraction2}
	There exists an unique Brenier potential $ u_{T}^{\gets} $ satisfies $ \nabla u_{T}^{\gets}=(\nabla u_{T}^{\to})^{-1} $ and Monge-Ampère equation \eqref{eq:Monge-Ampere_equation} with boundary constraints $ \rho_{0}=p_{\infty} $ and $ \rho_{1}=p_{T} $.
\end{remark}
The above analysis reveals that static OT is the optimal single-step method for prior error elimination: it not only achieves the minimal transport cost but also is intrinsically compatible with probability flow. Thus, we leverage this insight to accelerate DM generation.
\subsection{Applied to Accelerated Generation for DMs}
As established in Theorem \ref{theorem:OT_contraction}, static OT provides the most intrinsic accelerated sampler for DMs. Therefore, we choose a smaller time $ \varepsilon<T'\ll T $ to terminate the diffusion, and then replace the remaining complicated two-stage calculation of time-dependent OT in DMs with solving the Monge-Ampère equation \eqref{eq:Monge-Ampere_equation}, where $\rho_{0}=p_{\infty}$ and $\rho_{1}=p_{T'}$. Denoting the learned parameterized Brenier potential energy we learn is $ u_{\boldsymbol{h}}^{\gets} $, then our accelerated generation distribution can be concretized as $q_{\varepsilon}=M_{\boldsymbol{\theta}}^{T',\varepsilon}\circ\nabla u_{\boldsymbol{h}}^{\gets}(p_{\infty}) $.

Since $\nabla u_{\boldsymbol{h}}^{\gets}$ is a semi-discrete OT map, it suffers from poor generalization ability when directly applied to generation tasks. To improve generative diversity, we not only introduce stochasticity in diffusion sampling part of Algorithm \ref{alg:unconditional_generator}, but also achieve the continualization of $\nabla u_{\boldsymbol{h}}^{\gets}$ via a controllable smoothing technique. Specifically, given any smoothness coefficient $\tau>0$ we set $r_{i}=\frac{\langle \cdot,\boldsymbol{x}_{T'}^{i}\rangle +h_{i}}{\tau}$ and obtain the following controllable approximation of the Brenier potential $u_{\boldsymbol{h}}^{\gets}$ refer to \cite{mazumder2019computational,mustafi2021convex}\vspace{-10pt}

{\setlength{\abovedisplayskip}{0.15pt}
\setlength{\belowdisplayskip}{1pt}
\small
\begin{equation}
	\widetilde{u}_{\boldsymbol{h}}^{\gets}(\tau,\cdot)=\tau\log\left(\sum_{i\in\mathcal{I}}\exp\left(r_{i}\right)\right)-\tau\log|\mathcal{I}|,
\end{equation}
}\vspace{-10pt}

which satisfies error bound $\|u_{\boldsymbol{h}}^{\gets}-\widetilde{u}_{\boldsymbol{h}}^{\gets}\|_{\infty}\leq\tau\log|\mathcal{I}|$. The gradient of $\widetilde{u}_{\boldsymbol{h}}^{\gets}$ is an continuous approximation of $\nabla u_{\boldsymbol{h}}^{\gets}$\vspace{-10pt}

{\setlength{\abovedisplayskip}{0.15pt}
\setlength{\belowdisplayskip}{1pt}
\small
\begin{equation}\label{eq:Mot_smooth}
	\nabla \widetilde{u}_{\boldsymbol{h}}^{\gets}(\tau,\cdot)=\frac{\sum_{i\in\mathcal{I}}\exp\left(r_{i}\right)\boldsymbol{x}_{T'}^{i}}{\sum_{i\in\mathcal{I}}\exp\left(c_{i}\right)}.
\end{equation}
}\vspace{-10pt}

Thus, our sampling process is summarized in Algorithm \ref{alg:unconditional_generator}.
\begin{algorithm}[htbp]
	\footnotesize
	\caption{{\small Sampling Process with OT Accelerator}}
	\label{alg:unconditional_generator}
	\textbf{Input}: Initial distribution $ p_{\infty} $, smoothed OT $ \nabla \widetilde{u}_{\boldsymbol{h}}^{\gets} $ with smoothness coefficient $\tau$, trained score network $ \boldsymbol{S}_{\boldsymbol{\theta}}(\boldsymbol{x},t) $ over $ [\varepsilon,T'] $, discrete time interval $ \Delta t $ such that $T'-\varepsilon=m\Delta t$.
	
	\begin{algorithmic}[1]
		\STATE $\boldsymbol{z}\sim p_{\infty}$.
		\STATE $\boldsymbol{x}_{T'}\gets\nabla \widetilde{u}_{\boldsymbol{h}}^{\gets}(\tau,\boldsymbol{z})$.
		\FOR{$ i=0:m-1 $}
		\STATE $ \boldsymbol{x}_{t_{i+1}}\gets \boldsymbol{x}_{t_{i}}+(f(t_{i})\boldsymbol{x}_{t_{i}}-g(t_{i})^{2}\boldsymbol{S}_{\boldsymbol{\theta}}(\boldsymbol{x}_{t_{i}},t_{i}))\Delta t+g(t_{i})\cdot\sqrt{\Delta t}\boldsymbol{\xi}$, where $t_{i}= T'-i\Delta t,\boldsymbol{\xi}\sim\mathcal{N}(\boldsymbol{0},\boldsymbol{I})$.
		\ENDFOR
	\end{algorithmic}
	\textbf{Return}: Generated sample $ \boldsymbol{x}_{\varepsilon} $.
\end{algorithm}
\begin{table}[htbp]
	\caption{Comparison of prior error $\mathcal{W}_{2}(p_{T},q_{T})$ elimination effects among different generators. We set $T = 500$ and $q_{T}=\mathcal{G}(p_{\infty})$. Additionally, Id denotes the identity map, ie. $p_{\infty}=\text{Id}(p_{\infty})$. AM: Anchored map, FB: Flow-based, PS: Prometheus SDEs.}\label{tab:Comparison of Prior Error Elimination Effects}
	\centering
	\scalebox{0.772}{
	\begin{tabular}{c|cccccccc}
		\hline
		$\mathcal{G}(\cdot)$ & Id & GAN & VAE & AM & FB & AE-OT & PS & $\nabla u_{\boldsymbol{h}}^{\gets}$\\ \hline
		Cifar10 & 0.19 & 0.03 & 0.04 & 0.04 & 0.05 & 0.02 & 0.06 & \textbf{0.00}\\
		CelebA & 0.36 & 0.10 & 0.10 & 0.12 & 0.09 & 0.04 & 0.20 & \textbf{0.00}\\
		FFHQ & 0.70 & 0.15 & 0.17 & 0.21 & 0.23 & 0.10 & 0.31 & \textbf{0.00}\\
		AFHQ & 1.04 & 0.28 & 0.29 & 0.25 & 0.30 & 0.11 & 0.33 & \textbf{0.00}
		\\\hline
	\end{tabular}}
	\scalebox{0.730}{
	\begin{tabular}{c|cccccccc}
		\hline
		$\mathcal{G}(\cdot)$ & $ \nabla \widetilde{u}_{\boldsymbol{h}}^{\gets} $ & $ \nabla \widetilde{u}_{\boldsymbol{h}}^{\gets} $ & $ \nabla \widetilde{u}_{\boldsymbol{h}}^{\gets} $ & $ \nabla \widetilde{u}_{\boldsymbol{h}}^{\gets} $ & $ \nabla \widetilde{u}_{\boldsymbol{h}}^{\gets} $ & $ \nabla \widetilde{u}_{\boldsymbol{h}}^{\gets} $ & $ \nabla \widetilde{u}_{\boldsymbol{h}}^{\gets} $ & $ \nabla \widetilde{u}_{\boldsymbol{h}}^{\gets} $\\ \hline
		$\tau$ & 0.01 & 0.02 & 0.03 & 0.04 & 0.05 & 0.06 & 0.07 & 0.08\\ \hline
		Cifar10 & \underline{0.01} & 0.02 & 0.02 & 0.05 & 0.06 & 0.08 & 0.10 & 0.12\\
		CelebA & \underline{0.03} & 0.05 & 0.08 & 0.12 & 0.17 & 0.21 & 0.25 & 0.30\\
		FFHQ & \underline{0.08} & 0.10 & 0.13 & 0.17 & 0.22 & 0.30 & 0.39 & 0.44\\
		AFHQ & \underline{0.09} & 0.12 & 0.16 & 0.20 & 0.31 & 0.43 & 0.62 & 0.81
		\\\hline
	\end{tabular}}
	\label{tab:Comparison of prior error elimination effects among different single-step generators}
\end{table}
\begin{figure}[t]
	\centering
	\renewcommand{\figurename}{Fig.}
	\renewcommand{\thefigure}{\arabic{figure}}
	\includegraphics[width=0.484\textwidth]{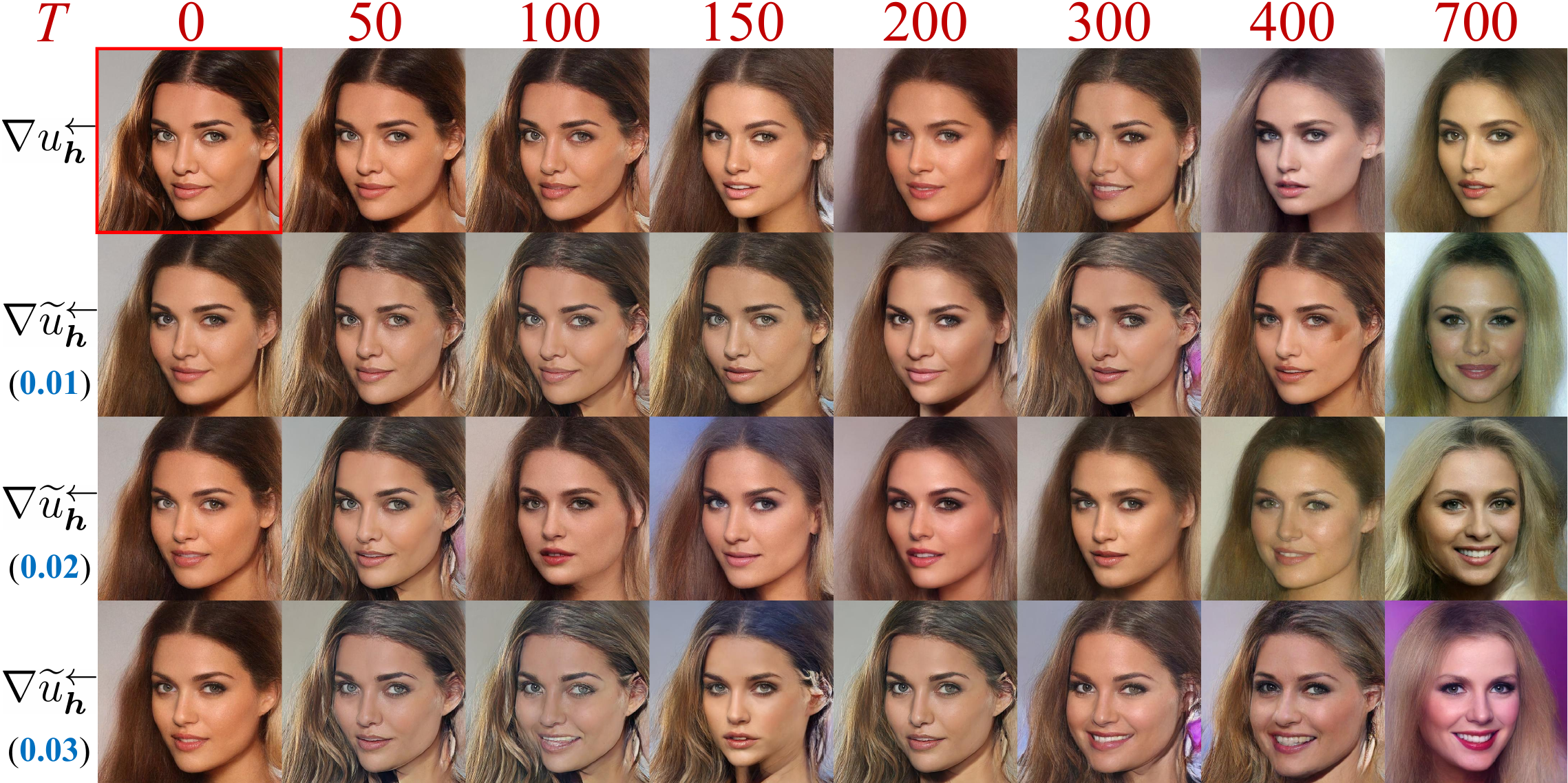}
	\caption{Generalization brought by $T$ and $\tau$. \textcolor{red}{Red}: original image.}
	\label{fig:T_tau}
\end{figure}

Fig. \ref{fig:T_tau} visualizes the effects of varying $\tau$ and diffusion time $T$ on model generalization, while Tab. \ref{tab:Comparison of Prior Error Elimination Effects} presents the prior error elimination efficacy of various $\mathcal{G}(\cdot)$. Results show that OT’s geometric properties enable full prior error elimination, aligning with Lemma \ref{lemma:eliminate the prior error}. In contrast, other $\mathcal{G}(\cdot)$ are continuous maps and only mitigate prior error. Notably, OT’s geometric characteristics weaken with smoothing techniques. Thus, we set $\tau=0.02$ for all subsequent experiments to balance prior error elimination and generalization.
\subsection{Extended Discussion}
We include this section in Appendix B, which comprises subsections (B.1) \textit{Why Terminate Diffusion at $T’$ Instead of $\varepsilon$?}, (B.2) \textit{Relationship with ODE-based Accelerated Sampling for DMs}, (B.3) \textit{Core Differences from Works \cite{li2023dpm, li2025optimal}} and (B.4) \textit{The Advantages/Disadvantages of Static and Dynamic Optimal Transport}.
\section{Experiments}\label{sec4}
\subsection{Experiment Settings}
\textbf{Datasets.} We mainly use four public image datasets, consisting of Cifar10 32$\times$32 \cite{krizhevsky2009learning}, CelebA 64$\times$64 \cite{liu2015deep} and FFHQ 256$\times$256 \cite{ImageNetrussakovsky2015imagenet}, AFHQ 512$\times$512 \cite{choi2020stargan}.

\textbf{Baseline Methods.} We categorize the comparative methods into three types, all of which are elaborated in Tab. 6: \textit{(1) GAN-based methods.} \textit{(2) classical DMs and their variants.} \textit{(3) truncation and ODE-based acceleration strategies.}

\textbf{Evaluation Metrics.} We take the diffusion terminal time $T$ and the Number of Function Evaluations (NFE) as modeling efficiency criteria, use the prior error $\mathcal{W}_{2}(p_{T},q_{T})$ to measure the effects of various eliminators, and adopt the Fréchet Inception Distance (FID) \cite{heusel2017gans}, Mode Mixture Ratio (MMR) \cite{li2023dpm} and Precision \& Recall \cite{kynkaanniemi2019improved} as generation quality metrics.

Comprehensive experimental parameters and detailed visual results are provided in Appendix G.
\begin{figure}[b]
	\centering
	\renewcommand{\figurename}{Fig.}
	\renewcommand{\thefigure}{\arabic{figure}}
	\includegraphics[width=0.484\textwidth]{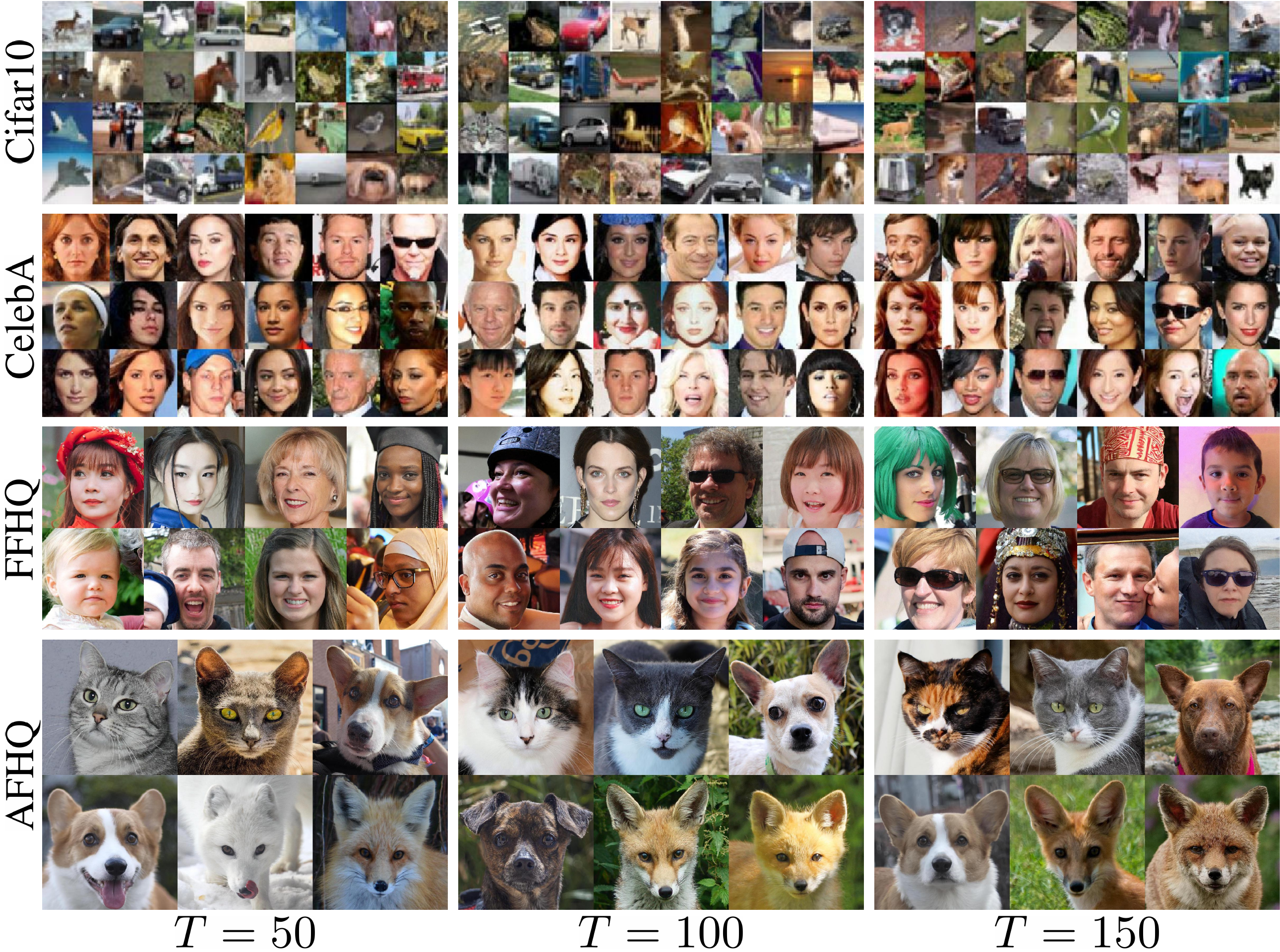}
	\caption{Generated images on four datasets with OT accelerator.}
	\label{fig:Generated images on four datasets with OT accelerator}
\end{figure}
\subsection{Evaluation of Prior Error and Sampling Efficiency}
\begin{table}[htbp]
	\caption{Quantitative comparison of different models across four datasets. $p_{T}$ and $q_{T}=\mathcal{G}(\mathcal{N}(\boldsymbol{0},\boldsymbol{I}))$ imply the forward termination and reverse initial distribution, respectively. Best performance is in \textbf{bold}, suboptimal in \underline{underline}. Additionally, $\downarrow$/$\uparrow$ mark the desired value direction. \textcolor{gray}{Gray backgrounds} denote GAN-based models; others represent classical DMs and their variants. Their optimal values are marked separately. All baselines are listed in Tab. 6. Additionally, Id is the identity map, ie. $\mathcal{N}(\boldsymbol{0},\boldsymbol{I})=\text{Id}(\mathcal{N}(\boldsymbol{0},\boldsymbol{I}))$.}\label{tab:Quantitative comparison}
	\centering
	\scalebox{0.59}{
		\begin{tabular}{c|ccc|cccc}
			\hline
			Methods & $T$ & $\mathcal{G}(\cdot)$ & $\mathcal{W}_{2}(p_{T},q_{T})\downarrow$ & NFE$\downarrow$ & FID$\downarrow$ & Precision$\uparrow$ & Recall$\uparrow$ \\ \hline
			\hline
			\rowcolor[HTML]{96FFFB}\multicolumn{8}{c}{Cifar10 32$\times$32}\\ \hline
			\rowcolor[HTML]{C0C0C0}ViTGAN & --- & --- & --- & 1 & 4.57 & --- & --- \\
			\rowcolor[HTML]{C0C0C0}Styleformer & --- & --- & --- & 1 & \underline{2.82} & --- & --- \\
			\rowcolor[HTML]{C0C0C0}StyleGAN-XL & --- & --- & --- & 1 & \textbf{1.85} & --- & --- \\ \hline
			DDIM & 1000 & Id & 0.0087 & 100 & 4.60 & --- & --- \\
			DDPM & 1000 & Id & 0.0087 & 1000 & 3.31 & \textbf{0.70} & 0.59 \\
			LSGM & --- & --- & --- & 23 & 2.01 & 0.65 & 0.58 \\
			PD & --- & --- & --- & 2 & 4.51 & --- & --- \\
			WaveDiff & --- & --- & --- & 4 & 4.01 & 0.61 & 0.54 \\
			EDM & --- & --- & --- & 27 & 3.73 & 0.63 & 0.56 \\
			CM & --- & --- &---  & 2 & 2.93 & --- & --- \\
			2-RectFlow & --- & --- & --- & 110 & 3.26 & 0.57 & 0.54\\
			1-RectFlow & --- & --- & --- & 127 & 2.58 & 0.59 & 0.60 \\
			ECT & --- & --- & --- & 2 & 1.94 & --- & --- \\ 
			SiD & --- & --- & --- & 1 & 1.81 & --- & 0.63 \\ 
			EDM-AOT & --- & --- & --- & 29 & 1.73 & 0.65 & 0.61 \\
			GDD & --- & --- & --- & 1 & 1.66 & --- & --- \\ \hline
			DPM-Solver & 1000 & Id & 0.0087 & 50 & 2.65 &  ---& --- \\
			DPM-Solver & 1000 & Id & 0.0087 & 20 & 3.72 & 0.58 & 0.53 \\
			DEIS & 1000 & Id & 0.0087 & 20 & 2.86 & 0.61 & 0.67 \\
			UciPC & 1000 & Id & 0.0087 & 10 & 3.84 & 0.64 & 0.70 \\
			GENIE & 1000 & Id & 0.0087 & 20 & 3.94 & --- & --- \\ \hline
			TDPM & 100 & GANs & 0.0812 & 100 & 2.97 & --- & 0.57 \\
			TDPM+ & 100 & GANs & 0.0812 & 100 & 2.83 & --- & 0.58 \\
			ES-DDPM & 700 & GANs & 0.0171 & 700 & 3.11 & 0.59 & 0.63 \\
			ES-DDPM & 200 & GANs & 0.0527 & 200 & 5.02 & 0.62 & 0.65 \\
			DiNof & 600 & FB & 0.0342 & 600 & 2.01 & 0.58 & 0.68 \\
			OTGD & 150 & AE-OT & 0.0245 & 150 & 2.90 & --- & 0.71 \\
			TCM & 80 & CMs & --- & 2 & 2.46 & --- & --- \\ \hline
			Ours & 100 & $ \nabla u_{\boldsymbol{h}}^{\gets} $ & \textbf{0.0000} & 100 & 1.96 & 0.65 & 0.68 \\
			Ours & 50 & $ \nabla u_{\boldsymbol{h}}^{\gets} $ & \textbf{0.0000} & 50 & \textbf{1.31} & \underline{0.66} & \textbf{0.76} \\
			Ours ($\tau=0.02$) & 100 & $ \nabla \widetilde{u}_{\boldsymbol{h}}^{\gets} $ & 0.0242 & 100 & 2.00 & 0.60 & 0.66 \\
			Ours ($\tau=0.02$) & 50 & $\nabla \widetilde{u}_{\boldsymbol{h}}^{\gets} $ & 0.0457 & 50 & \underline{1.34} & 0.63 & \underline{0.72}\\
			\hline
			\hline
			\rowcolor[HTML]{96FFFB}\multicolumn{8}{c}{CelebA 64$\times$64}\\ \hline
			\rowcolor[HTML]{C0C0C0}ViTGAN & --- & --- & --- & 1 & 3.74 & --- & --- \\
			\rowcolor[HTML]{C0C0C0}Styleformer & --- & --- & --- & 1 & 3.92 & --- & --- \\
			\rowcolor[HTML]{C0C0C0}LadaGAN & --- & --- & --- & 1 & \underline{1.81} & --- & --- \\
			\rowcolor[HTML]{C0C0C0}D-StyleGAN2 & --- & --- & --- & 1 & \textbf{1.69} & --- & --- \\ \hline
			DDIM & 1000 & Id & 0.0283 & 200 & 6.53 & 0.75 & 0.42 \\
			DDPM & 1000 & Id & 0.0283 & 1000 & 3.26 & 0.65 & 0.63\\
			PDM & --- & --- & --- & 50 & 1.77 & --- & ---\\
			EDM & --- & --- & --- & 50 & \underline{1.66} & --- & ---\\
			ADM-IP & 1000 & --- & --- & 200 & \textbf{1.53} & 0.72 & 0.68\\
			EDDPM & 1000 & --- & --- & 50 & 6.65 & 0.64 & 0.59\\ \hline
			DPM-Solver & 1000 & Id & 0.0283 & 20 & 3.13 & 0.71 & 0.46\\
			DEIS & 1000 & Id & 0.0283 & 50 & 2.95 & 0.65 & 0.62\\ \hline
			TDPM+ & 50 & GANs & 0.0931 & 50 & 3.28 & 0.61 & 0.62 \\
			ES-DDPM & 200 & GANs & 0.0476 & 200 & 2.55 & 0.60 & 0.58 \\
			OTGD & 150 & AE-OT & 0.0470 & 150 & 1.87 & --- & 0.61 \\ \hline
			Ours & 100 & $ \nabla u_{\boldsymbol{h}}^{\gets} $ & \textbf{0.0000} & 100 & 2.15 & \underline{0.77} & 0.69 \\
			Ours & 50 & $ \nabla u_{\boldsymbol{h}}^{\gets} $ & \textbf{0.0000} & 50 & 1.88 & \textbf{0.79} & \textbf{0.75} \\
			Ours ($\tau=0.02$) & 100 & $ \nabla \widetilde{u}_{\boldsymbol{h}}^{\gets} $ & 0.0531 & 100 & 2.17 & 0.73 & 0.65 \\
			Ours ($\tau=0.02$) & 50 & $\nabla \widetilde{u}_{\boldsymbol{h}}^{\gets} $ & 0.0587 & 50 & 1.92 & 0.75 & \underline{0.72}\\
			\hline
			\hline
			\rowcolor[HTML]{96FFFB}\multicolumn{8}{c}{FFHQ 256$\times$256}\\ \hline
			\rowcolor[HTML]{C0C0C0}StyleGAN-XL & --- & --- & --- & 1 & \underline{2.19} & --- & ---\\
			\rowcolor[HTML]{C0C0C0}StyleSAN-XL & --- & --- & --- & 1 & \textbf{1.68} & --- & ---\\ \hline
			ADM-G & 1000 & --- & --- & 250 & 3.94 & 0.67 & 0.66\\
			EDM & --- & --- & --- & 79 & 2.39 & 0.69 & 0.68\\
			DiT-L/2 & --- & --- & --- & 88 & 4.55 & --- & ---\\ \hline
			DPM-Solver & 1000 & Id & 0.0920 & 50 & 7.39 & 0.72 & 0.66\\
			UciPC & 1000 & Id & 0.0920 & 10 & 6.99 & 0.70 & \underline{0.71}\\ \hline
			Ours & 100 & $ \nabla u_{\boldsymbol{h}}^{\gets} $ & \textbf{0.0000} & 100 & 2.56 & \underline{0.76} & 0.70 \\
			Ours & 50 & $ \nabla u_{\boldsymbol{h}}^{\gets} $ & \textbf{0.0000} & 50 & \textbf{2.14} & \textbf{0.77} & \textbf{0.72} \\
			Ours ($\tau=0.02$) & 100 & $ \nabla \widetilde{u}_{\boldsymbol{h}}^{\gets} $ & 0.1083 & 100 & 2.84 & 0.71 & 0.67 \\
			Ours ($\tau=0.02$) & 50 & $\nabla \widetilde{u}_{\boldsymbol{h}}^{\gets} $ & 0.1179 & 50 & \underline{2.17} & 0.75 & 0.70\\
			\hline
			\hline
			\rowcolor[HTML]{96FFFB}\multicolumn{5}{c}{\multirow{1}{*}{~~~~~~~~~~~~~~~~~~~~~~~~~~~~~~~~~~~~~~~~AFHQ 512$\times$512}}&\multicolumn{1}{|c}{}&FID$\downarrow$&\\ \hline
			Methods& $T$ & $\mathcal{G}(\cdot)$ & $\mathcal{W}_{2}(p_{T},q_{T})\downarrow$ & NFE$\downarrow$ & \multicolumn{1}{|c}{Cat} & Dog & Wild \\ \hline
			\rowcolor[HTML]{C0C0C0}StyleGAN2-ADA& --- & --- & --- & 1 & \multicolumn{1}{|c}{3.55} & 7.41 &3.05 \\
			\rowcolor[HTML]{C0C0C0}Projected GAN& --- & --- & --- & 1 & \multicolumn{1}{|c}{\textbf{2.16}} & \textbf{4.52} & \underline{2.17}\\
			\rowcolor[HTML]{C0C0C0}Diffusion InsGen& --- & --- & --- & 1 & \multicolumn{1}{|c}{\underline{2.40}} & 4.83 & \textbf{1.51}\\
			\rowcolor[HTML]{C0C0C0}Vision-aided GAN & --- & --- & --- & 1 & \multicolumn{1}{|c}{2.53} & \underline{4.73} & 2.36\\ \hline
			GENIE& 1000 & Id & 0.1751 & 15 & \multicolumn{1}{|c}{4.83} & --- & ---\\
			DDMI & --- & --- & --- & 50 & \multicolumn{1}{|c}{4.27} & 8.54 & ---\\ \hline
			Ours & 100 & $\nabla u_{\boldsymbol{h}}^{\gets} $ & \textbf{0.0000} & 100 & \multicolumn{1}{|c}{2.91} & 2.88 & 1.76\\
			Ours & 50 & $\nabla u_{\boldsymbol{h}}^{\gets} $ & \textbf{0.0000} & 50 & \multicolumn{1}{|c}{\textbf{2.24}} & \textbf{2.35} & \textbf{1.34}\\
			Ours ($\tau=0.02$) & 100 & $\nabla \widetilde{u}_{\boldsymbol{h}}^{\gets} $ & 0.1475 & 100 & \multicolumn{1}{|c}{3.03} & 2.99 & 1.87\\
			Ours ($\tau=0.02$) & 50 & $\nabla \widetilde{u}_{\boldsymbol{h}}^{\gets} $ & 0.1513 & 50 & \multicolumn{1}{|c}{\underline{2.28}} & \underline{2.42} & \underline{1.43}\\ \hline
	\end{tabular}}
\end{table}
We collect the terminal time $T$ for all diffusion-based baselines without noise scale adjustment, and compute prior error $\mathcal{W}_{2}(p_{T},q_{T})$ using the POT library \cite{flamary2021pot}. Results from Fig. \ref{fig:W2_forward_reverse} and Tab. \ref{tab:Quantitative comparison} demonstrate that truncated acceleration methods can achieve relatively low prior errors without requiring excessively long $T$, in comparison with classical DMs and ODE-based acceleration methods. This is because the reverse initial distribution for the latter is typically set to $q_{T}=\mathcal{N}(\boldsymbol{0},\boldsymbol{I})$, thereby neglecting the impact of the prior error. In contrast, the former adopts the distribution $q_{T}=\mathcal{G}(\mathcal{N}(\boldsymbol{0},\boldsymbol{I}))$, which can effectively mitigate the prior error. We also compare the elimination effectiveness of different $\mathcal{G}(\cdot)$ at $T=500$ in Tab. \ref{tab:Comparison of Prior Error Elimination Effects}, showing that our method gives the optimal performance. Furthermore, Tab. \ref{tab:Quantitative comparison} reveals a key insight: regardless of the terminal time, unsmoothed OT can fully eliminate the prior error, which aligns with the theoretical claim of this paper (Lemma \ref{lemma:eliminate the prior error}).

\subsection{Quantification of Generation Quality}
It can be observed from the last three columns of Tab. \ref{tab:Quantitative comparison} that the proposed model outperforms or is comparable to all diffusion-based baseline models in terms of FID, Precision and Recall across all image datasets (50 NFEs). Fig. \ref{fig:Generated images on four datasets with OT accelerator} illustrates the generation results of our method on four datasets.
\begin{figure}[t]
	\setlength{\abovecaptionskip}{0cm}
	\setcounter{subfigure}{0}
	\centering
	\begin{subfigure}[b]{0.484\textwidth}
		\includegraphics[width = 0.989\textwidth]{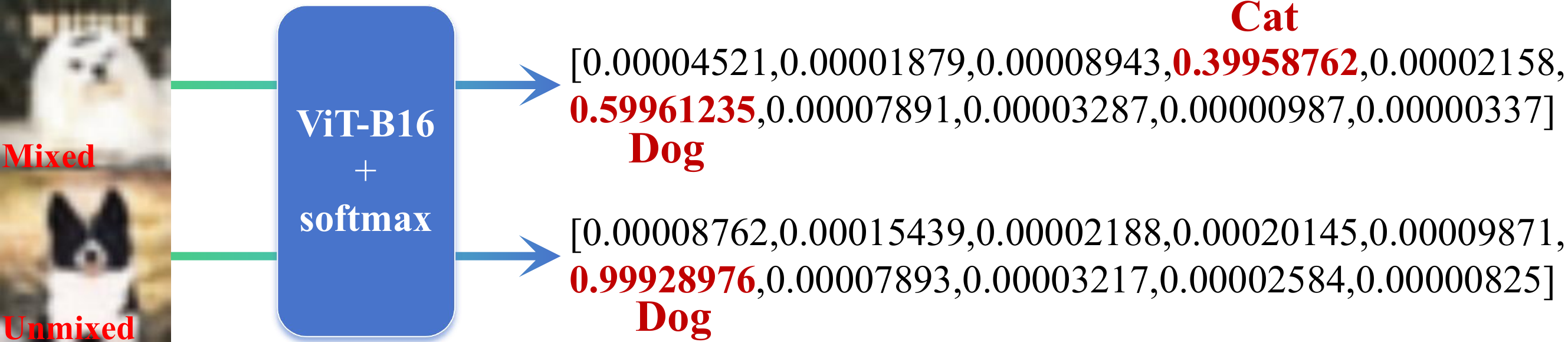}
		\caption{Using ViT-B16 to Distinguish Mixed and Unmixed Images.}
		\label{fig:mode_mixture_test:a}
	\end{subfigure}
	\\
	\begin{subfigure}[b]{0.484\textwidth}
		\includegraphics[width = 0.989\textwidth]{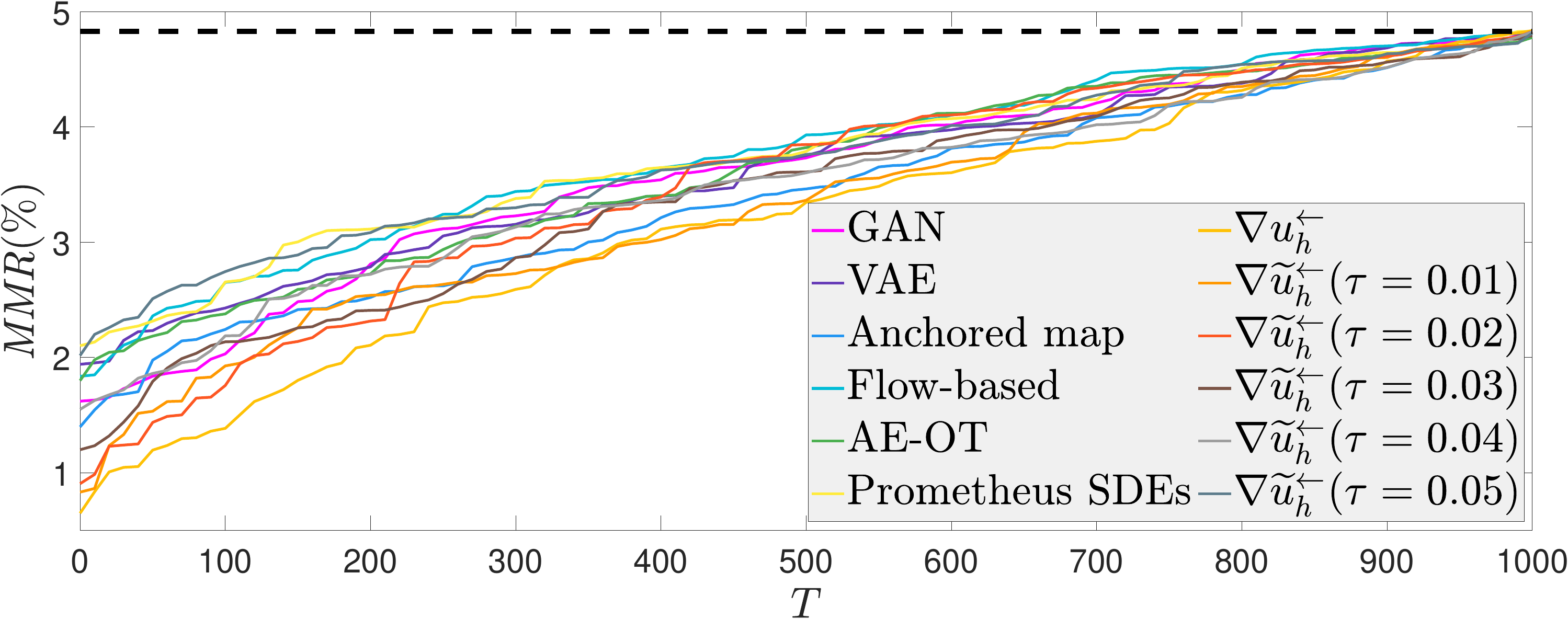}
		\caption{MMR Comparison for Different $\mathcal{G}(\cdot)$-Truncations on Cifar10.}
		\label{fig:mode_mixture_test:b}
	\end{subfigure}
	\caption{MMR in \eqref{eq:MMR} quantifies the degree of mode mixture, based on that mixed and unmixed images exhibit distinct behaviors under a classifier. Subfigure \textit{(b)} illustrates the variation of MMR ($\lambda=0.2$) with respect to $T$ for different truncated models. A lower MMR indicates that generated images are more class-discriminative and well-separated.}\label{fig:mode_mixture_test}\vspace{-10pt}
\end{figure}

\vspace{-13pt}
For generated image set $\{\boldsymbol{x}_{\varepsilon}^{i}\}_{i=1}^{N_{\textit{mmr}}}$, we adopt the industry-recognized SOTA pre-trained classifier ViT-B16 \cite{dosovitskiy2020image} to estimate the class membership probability of each generated image. Typically, the class probability vector of a mixed image contains more than two components exceeding the predefined threshold $\lambda$, whereas that of an unmixed image exhibits a single dominant component (Fig. \ref{fig:mode_mixture_test:a}). Hence, the MMR is defined as follows \cite{li2023dpm}\vspace{-10pt}

{\setlength{\abovedisplayskip}{0.15pt}
\setlength{\belowdisplayskip}{1pt}
\small
\begin{equation}\label{eq:MMR}
	\text{MMR}=\frac{1}{N_{\textit{mmr}}}\sum_{i=1}^{N_{\textit{mmr}}}\mathbb{I}_{|\{p_{j}^{i}|p_{j}^{i}\geq\lambda,j=1,2,\cdots,C\}|\geq 2}(\boldsymbol{x}_{\varepsilon}^{i}),
\end{equation}
}\vspace{-10pt}

where $C$ denotes the number of classes, $p_{j}^{i}$ represents the probability that ViT-B16+softmax classifies sample $\boldsymbol{x}_{\varepsilon}^{i}$ as the $j$-th class, and $\mathbb{I}_{A}$ is the indicator function over set $A$.

Mode mixture is essentially a result of model overgeneralization, producing meaningless samples with mixed inter-class features. DMs' stochasticity and $\mathcal{G}(\cdot)$'s smoothness are key drivers of this phenomenon. Using different $\mathcal{G}(\cdot)$ as prior error eliminators on the same DDPM architecture, we computed their MMR values. Fig. \ref{fig:mode_mixture_test} shows that as $T$ increases, mode mixture worsens across all models, ultimately converging to vanilla DDPM’s MMR. Notably, unsmoothed OT yields the lowest MMR among all $\mathcal{G}(\cdot)$ by fully eliminating prior errors. In contrast, OT smoothing blurs inter-class boundaries, increasing MMR. This further confirms that our method regulates generalization to control generated image quality—a unique advantage absent in other $\mathcal{G}(\cdot)$.
\subsection{Ablation Study}
(1) Diffusion termination time $T$. Fig. \ref{fig:W2_forward_reverse:b} illustrates the variation of prior errors and the deviation between generated and true distributions with $T$ for both the classic DDPM and our method. Figs. \ref{fig:T_tau}, \ref{fig:Generated images on four datasets with OT accelerator} and Tab. \ref{tab:Quantitative comparison} demonstrate the effects of different $T$ on the generative generalization, visual performance and generative quality of our method, respectively, while Fig. \ref{fig:mode_mixture_test:b} shows the variation of MMR scores with $T$ under different truncation strategies.

(2) Generator $\mathcal{G}(\cdot)$. Fig. \ref{fig:W2_forward_reverse:b} shows that incorporating the prior error eliminator brings the generated distribution closer to the ground truth. Tab. \ref{tab:Comparison of Prior Error Elimination Effects} and Fig. \ref{fig:mode_mixture_test:b} present the impacts of different generators $\mathcal{G}(\cdot)$ on prior error elimination and MMR scores, respectively.

(3) Smoothness coefficient $\tau$. Tab. \ref{tab:Comparison of Prior Error Elimination Effects}, Fig. \ref{fig:T_tau} and Fig. \ref{fig:mode_mixture_test:b} compare the effects of different $\tau$ on prior error elimination, model generalization and mode mixture scores, respectively.
\section{Conclusion}\label{sec5}
We propose an OT-based prior error elimination framework: constructing an OT map between the reverse initial and forward terminal distributions to align them precisely; quantifying the error upper bound via the Wasserstein distance to prove effective prior error elimination; additionally, deriving the asymptotic consistency between dynamic OT and probability flow to verify compatibility with the diffusion process’s intrinsic mechanism. Experiments show our method completely eliminates prior errors (theoretically and practically), while enhancing generation quality and accelerating inference. Unlike existing heuristic strategies, it has theoretical rigor and generalization, offering a universal solution for DMs optimization. Future work will explore the in-depth OT-DMs correlation in generalized scenarios to expand their application in complex tasks.

\clearpage
\bibliography{example_paper}
\bibliographystyle{icml2026}

\end{document}